\documentclass[10pt,twocolumn,letterpaper]{article}

\usepackage{iccv}
\usepackage{times}
\usepackage{epsfig}
\usepackage{graphicx}
\usepackage{amsmath}
\usepackage{amssymb}
\usepackage{booktabs}
\usepackage{array,multirow}

\newcommand{\myparagraph}[1]{\vspace{0.1em}\noindent\textbf{#1}}

\usepackage[pagebackref=true,breaklinks=true,letterpaper=true,colorlinks,bookmarks=false]{hyperref}

\iccvfinalcopy 

\ificcvfinal\fi

\begin{document}
	
\title{ GNeRF: GAN-based Neural Radiance Field without Posed Camera }

\author{Quan Meng$^{1}$
\quad
Anpei Chen$^{1}$
\quad
Haimin Luo$^{1}$
\quad
Minye Wu$^{1}$\\
Hao Su$^{2}$
\quad
Lan Xu$^{1}$
\quad
Xuming He$^{1}$
\quad
Jingyi Yu$^{1}$\\
$^{1}$ Shanghai Engineering Research Center of Intelligent Vision and Imaging \\ School of Information Science and Technology, \\ ShanghaiTech University  \qquad $^{2}$ University of California, San Diego\\
{\tt\small {\{mengquan,chenap,luohm,wumy,xulan1,hexm,yujingyi\}}@shanghaitech.edu.cn \quad {\{haosu\}}@eng.ucsd.edu}
}
	
\maketitle
\ificcvfinal\thispagestyle{empty}\fi
	
\begin{abstract}
	We introduce GNeRF, a framework to marry Generative Adversarial Networks (GAN) with Neural Radiance Field (NeRF) reconstruction for the complex scenarios with unknown and even randomly initialized camera poses. Recent NeRF-based advances have gained popularity for remarkable realistic novel view synthesis. However, most of them heavily rely on accurate camera poses estimation, while few recent methods can only optimize the unknown camera poses in roughly forward-facing scenes with relatively short camera trajectories and require rough camera poses initialization. Differently, our GNeRF only utilizes randomly initialized poses for complex outside-in scenarios. We propose a novel two-phases end-to-end framework. The first phase takes the use of GANs into the new realm for optimizing coarse camera poses and radiance fields jointly, while the second phase refines them with additional photometric loss. We overcome local minima using a hybrid and iterative optimization scheme. Extensive experiments on a variety of synthetic and natural scenes demonstrate the effectiveness of GNeRF. More impressively, our approach outperforms the baselines favorably in those scenes with repeated patterns or even low textures that are regarded as extremely challenging before. 
\end{abstract}
	
\begin{figure}[tbp] 
	\centering 
	\includegraphics[width=1\linewidth]{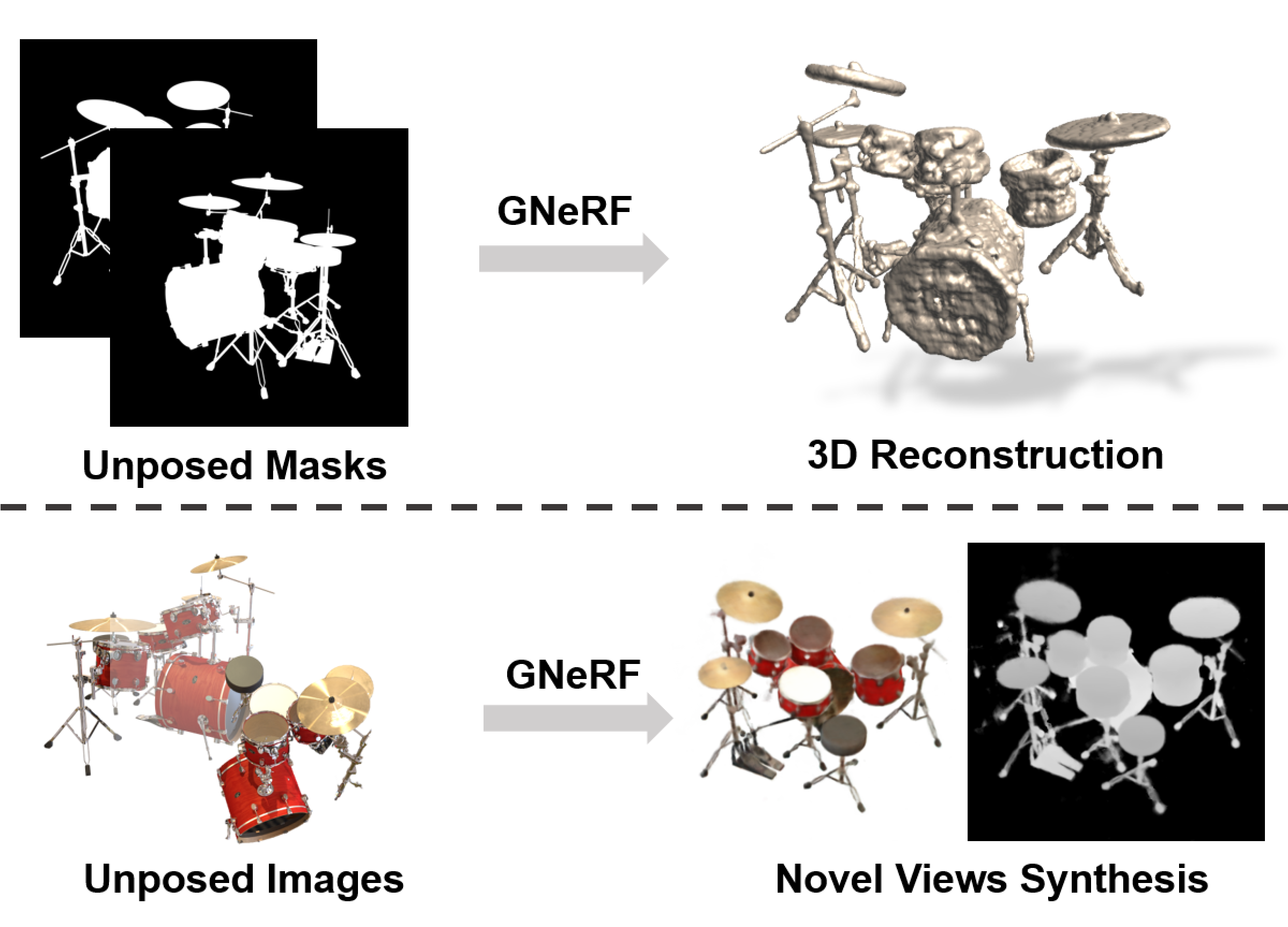} 
	\caption{Our approach estimates both camera poses and neural radiance fields using only randomly initialized poses in complex scenarios, even in the extreme case when the input views are only texture-less gray masks.} 
	\label{fig:fig_1_teaser} 
	\vspace{-8pt} 
\end{figure} 

\section{Introduction}
Recovering 3D representations from multi-view 2D images is one of the core tasks in computer vision. Recently, significant progress has been made with the emergence of neural radiance fields methods (e.g., NeRF~\cite{nerf}), which represents a scene as a continuous 5D function and uses volume rendering to synthesize new views. Although NeRF and its follow-ups~\cite{mvsnerf, liu2020neural, martin2020nerf, wang2021ibrnet, zhang2020nerf++} achieve an unprecedented level of fidelity on a range of challenging scenes, most of these methods rely heavily on knowing the accurate camera poses, which is yet a long-standing but challenging task. The conventional camera pose estimation process suffers in challenging scenes with repeated patterns, varying lighting, or few keypoints, and building on these methods adds additional uncertainty to the NeRF training process.

To explore the possibilities of alleviating the dependence on accurate camera pose information, recently, iNeRF~\cite{yen2020inerf} and NeRF$--$~\cite{wang2021nerf} attempt to optimize camera pose along with other parameters when training NeRF. While certain progress has been made, both of them can only optimize camera poses when relatively short camera trajectories with reasonable camera pose initialization are available. It is worth noting that, NeRF$--$ is limited to roughly forward-facing scenes, the focus of iNeRF is camera pose estimation but not radiance field estimation, and it assumes a trained NeRF which in turn requires known camera poses as supervision. When greater viewpoint uncertainty presents, camera poses estimation is extremely challenging and prone to falling into local minima.

To this end, we propose \textbf{GNeRF}, a novel algorithm that can estimate both camera poses and neural radiance fields when the cameras are initialized at random poses in complex scenarios. Our algorithm has two phases: the first phase gets coarse camera poses and radiance fields with adversarial training; the second phase refines them jointly with a photometric loss. Taking the use of Generative Adversarial Networks (GANs) into the realm of camera poses estimation, we extend the NeRF model to jointly optimize 3D representation and camera poses in complex scenes with large displacements. Instead of directly propagating the photometric loss back to the camera pose parameters, which is sensitive to challenging conditions (e.g., less texture and varying lighting) and apt to fall into local minima, we propose a hybrid and iterative optimization scheme. Our learning pipeline is fully differentiable and end-to-end trainable,  allowing our algorithm to perform well in the challenging scenes where COLMAP-based~\cite{schonberger2016structure} methods suffer from challenges such as repeated patterns, low textures, noise, even in the extreme cases when the input views are a collection of gray masks, as shown in Fig.~\ref{fig:fig_1_teaser}. In addition, our method can predict new poses of images belonging to the same scene through the trained inversion network without tedious per-scene pose estimation (e.g., COLMAP-like methods) or time-consuming gradient-based optimization (e.g., iNeRF and NeRF$--$).
We experiment with our GNeRF on a variety of synthetic and natural scenes. We demonstrate results on par with COLMAP-based NeRF methods in regular scenes; more impressively, our method outperforms the baselines in cases with less texture that are regarded as extremely challenging before. 

\section{Related Works}

\myparagraph{Neural 3D Representations}
Classic approaches largely rely on discrete representations such as meshes~\cite{gkioxari2019mesh}, voxel grids~\cite{choy20163d, tatarchenko2017octree, wu20153d}, point clouds~\cite{fan2017point}. Recent neural continuous implicit fields are gaining increasing popularity, due to their capability of representing a high level of details~\cite{mescheder2019occupancy, park2019deepsdf, peng2020convolutional}. But these methods need costly 3D annotations. To bridge the gap between 2D information and 3D representations, differential rendering tackles such integration for end-to-end optimization by obtaining useful gradients of the rendering process~\cite{zhang2021stnerf, liu2019learning, nerf, saito2019pifu, sun2021neural}. Liu~\etal\cite{liu2019learning} proposes the first usage of neural implicit surface representations in differentiable rendering. Mildenhall~\etal\cite{nerf} proposes differentiable volume rendering and achieves more view-consistent reconstructions of the scene. However, they all assume accurate camera poses as a prerequisite. 

Recently, several methods attempt to reduce dependence on precomputed camera poses. Adding noise to the ground-
truth camera poses, IDR~\cite{yariv2020multiview} produces accurate 3D surface reconstruction by simultaneously learning 3D representation and camera poses. Adding random offset to ground-truth camera poses, iNeRF~\cite{yen2020inerf} performs pose estimation by inverting a trained neural radiance field. Initializing camera poses to the identity matrix, NeRF$--$~\cite{wang2021nerf} demonstrates satisfactory novel view synthesis results in forward-facing scenes by optimizing camera parameters and radiance field
jointly. In contrast to these methods, our method does not depend on camera pose initialization and is not sensitive to challenging scenes with less texture and repeated patterns.

\myparagraph{Pose Estimation}
Traditional techniques typically rely on Structured-from-Motion (SfM)~\cite{andrew2001multiple, faugeras2001geometry, schonberger2016structure, wu2013towards} which extracts local descriptor (e.g., SIFT~\cite{lowe2004distinctive}), performs matching to find 2D-3D correspondence, estimates candidate poses, and then chooses the best pose hypothesis by RANSAC~\cite{fischler1981random}. Other retrieval-based methods~\cite{chum2007total, irschara2009structure, philbin2007object, sivic2003video} find images similar to the query image and establish the 2D-3D correspondence efficiently by matching the query image against the database images. Recently, deep learning-based methods attempt to regress the camera pose directly from 2D images without the need of tracking. PoseNet~\cite{kendall2015posenet} is the firstly end-to-end approach that adopts a modified truncated GoogleNet as pose regressor. Different architectures~\cite{naseer2017deep, walch2017image, wu2017delving} or pose losses~\cite{brahmbhatt2018geometry, kendall2017geometric} are utilized which lead to a significant improvement. Auxiliary tasks such learning relative pose
estimation~\cite{radwan2018vlocnet++, valada2018deep} or semantic segmentation~\cite{radwan2018vlocnet++} lead to a further improvement. For a better generalization of the network, hybrid pose learning methods shift the learning towards local or related problems: \cite{balntas2018relocnet, laskar2017camera} propose to regress the relative pose of a query image to the known poses based on image retrieval.

These learning-based methods require large labeled training data, SSV~\cite{mustikovela2020self} proposes to estimate viewpoints from unlabeled images via self-supervision. Although great progress has been made, it still needs abundant training images. Our method belongs to learning-based methods but is trained per scene in a self-supervised manner.

\myparagraph{3D-Aware Image Synthesis} 
Generative adversarial nets, or more generally the paradigm of adversarial learning, have led to significant progress in various image synthesis tasks~\cite{karras2019style, mirza2014conditional, shaham2019singan}. But these methods operate on 2D space of pixels, ignoring the 3d structure of our natural scene. 3D-aware image synthesis correlates 3D model with 2D images, enabling explicit modification of 3D model~\cite{chan2020pi, chen2020free, henzler2019escaping, nguyen2018rendernet, nguyen2019hologan, niemeyer2020giraffe, schwarz2020graf}. Earlier 3D-aware image synthesis methods like RenderNet~\cite{nguyen2018rendernet} introduce rendering convolutional networks with a projection unit that can render 2D images from 3D shapes. PLATONICGAN~\cite{henzler2019escaping} uses a voxel-based representation and a family of differentiable rendering layers to discover the 3D structure of an object from an unstructured collection of 2D images. HoloGAN~\cite{nguyen2019hologan} introduces deep voxels representation and learns it also without any 3D shapes supervision. For these methods, the combination of differentiable rendering layers and implicit 3D representation can lead to entangled latent variables and destroy multi-view consistency. The most recent and relevant to ours are GRAF~\cite{schwarz2020graf}, GIRAFFE~\cite{niemeyer2020giraffe} and pi-GAN~\cite{chan2020pi}, with the expressiveness of NeRF, these methods allow disentangled shape, appearance modification of the generated objects. 

However, these methods require abundant data and focus on simplistic objects (e.g., faces, cars) instead of photo-realistic and complex scenes. Conversely, our method can handle complex real scenes with limited data by learning a coarse generative network with limited data and refining it with photometric constraints.

\begin{figure*}[t]
	\begin{center}
		\includegraphics[width=1.0\linewidth]{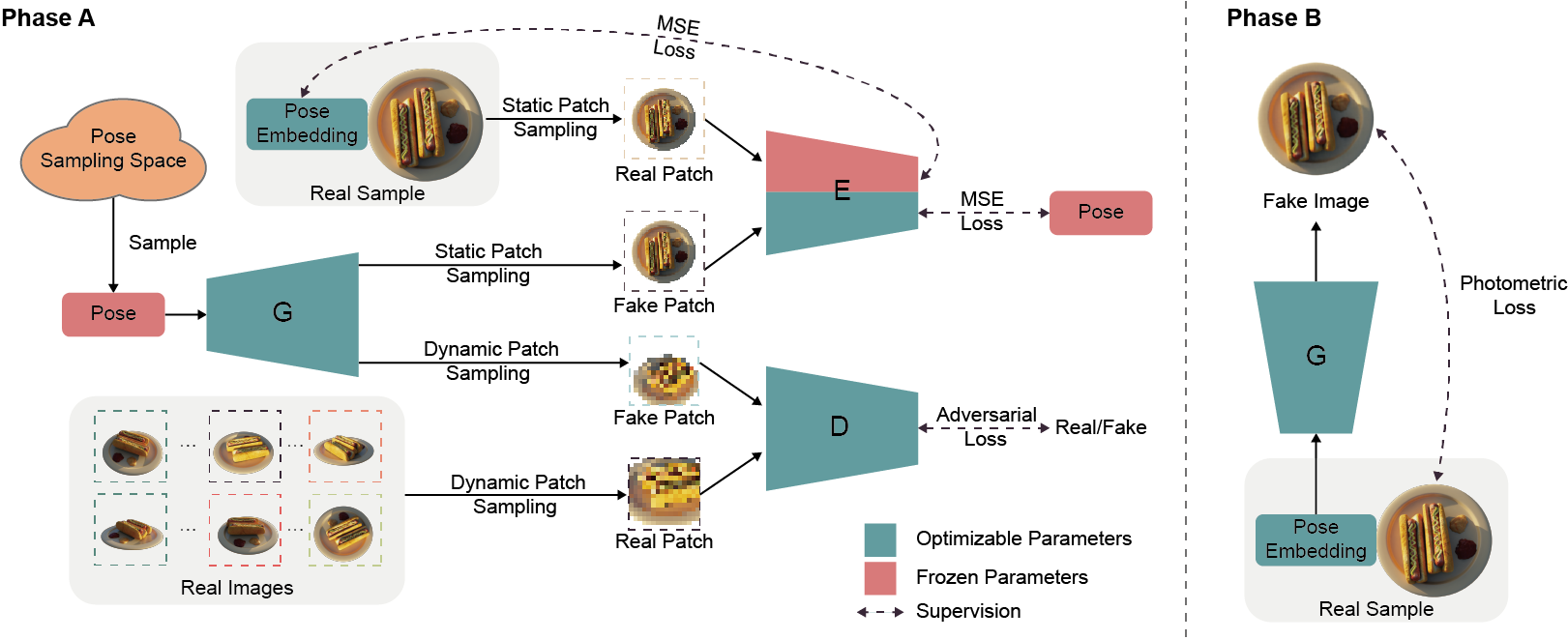}
	\end{center}
	\vspace{-0.2in}
	\caption{$\textbf{The pipeline of GNeRF.}$ Our pipeline learns the radiance fields and camera poses jointly in two phases. In phase A, we randomly sample poses from a predefined poses sampling space and generate corresponding images with the NeRF (G) model. The discriminator (D) learns to classify real and fake image patches. The inversion network (E) takes in the fake image patches and learns to output their poses. Then, with the inversion network's parameters frozen, we optimize the pose embeddings of real images in the dataset. In phase B, we utilize the photometric loss to refine radiance fields and pose embeddings jointly. We follow a hybrid and iterative optimization strategy of the pattern `A $\rightarrow$ AB$\dots$AB $\rightarrow$ B' in the training process. }
	\label{fig:pipeline}
\end{figure*}

\section{Preliminary}
We first introduce the basic camera and scene representation, as well as notations for our method in this section. 
\vspace{-7mm}
\paragraph{Camera Pose} Formally, we represent the camera pose/extrinsic parameters based on its position/location in 3D space and its rotation from a canonical view. For the camera position, we simply adopt a 3D embedding vector in Euclidean space, denoted as $\mathbf{t} \in \mathbb{R}^3$. For the camera rotation, the widely-used representations such as quaternions and Euler angles are discontinuous and difficult for neural networks to learn. Following the seminal work~\cite{zhou2019continuity}, we use a continuous 6D embedding vector $\mathbf{r} \in \mathbb{R}^6$ to represent 3D rotations, which is more suitable for learning. Concretely, given a rotation matrix $\mathbf{R} = \begin{bmatrix}\mathbf{a}_1 & \mathbf{a}_2 & \mathbf{a}_3\end{bmatrix}\in \mathbb{R}^{3\times 3}$, we compute the rotation vector $\mathbf{r}$ by dropping the last column of the rotation matrix. 

From the 6D pose embedding vector, we can also recover the original rotation matrix using a Gram-Schmidt-like process, in which the last column is computed by a generalization of the cross product to three dimension~\cite{zhou2019continuity}. 
\vspace{-3mm}
\paragraph{NeRF Scene Representation} We adopt the NeRF~\cite{nerf} framework to represent the underlying 3D scene and image formation, which encodes a scene as continuous volumetric radiance field of color and density. Specifically, given a 3D location $\mathbf{x}\in \mathbb{R}^3$ and 2D viewing direction $\mathbf{d} \in [-\pi,\pi]^2$ as inputs, the NeRF model defines a 5D vector-valued function $F_{\Theta}:(\mathbf{x}, \mathbf{d}) \rightarrow(\mathbf{c}, \sigma)$ based on an MLP network, where its outputs are an emitted color $\mathbf{c} \in \mathbb{R}^3$ and volume density $\sigma$, and $\Theta$ are network parameters. 
To render an image from a NeRF model, the NeRF model follows the classical volume rendering principles~\cite{kajiya1984ray}. 

For each scene, the NeRF framework learns a separate neural representation network with a dataset of RGB images of the scene, the corresponding camera poses and intrinsic parameters, and scene bounds. Concretely, given a dataset of calibrated RGB images $\mathcal{I} = \{I_1, I_2, \cdots, I_n\}$ of a single scene, the corresponding camera poses $\Phi = \{\phi_1, \phi_2, \cdots, \phi_n\}$ and a differentiable volume renderer $G$, the NeRF model optimizes the continuous volumetric scene function $F_{\Theta}$ by a photometric loss as below,
\begin{align}
	  \mathcal{L}_{N}(\Theta,\Phi) = \frac{1}{n}\sum_{i=1}^{n} \|I_i - \hat{I}_i\|^2_2, \quad 
	  \hat{I}_i = G(\phi_i;F_{\Theta})  
	  \label{nerf}
\end{align}

\section{Methods}

Our goal is to learn a NeRF model $F_{\Theta}$ from $n$ uncalibrated images $\mathcal{I}$ of a single scene without knowing their camera poses. To this end, we treat the camera poses $\Phi$ of those images as values of a latent variable, and propose an iterative learning strategy that jointly estimates the camera poses and learns the NeRF model. As the overview of our approach in Fig.~\ref{fig:pipeline} illustrates, the key ingredient of our method is a novel NeRF estimation strategy based on an integration of an adversarial loss and an inversion network (Phase A). This enables us to generate a coarse estimate of the implicit scene representation $F_{\Theta}$ and the camera poses $\Phi$ from a learned inversion network. Given the initial estimate, we utilize photometric loss to refine the NeRF scene model and those camera poses (Phase B). Interestingly, our pose-free NeRF estimation process can also further improve the refined scene representation and camera poses. Additionally, we develop a regularized NeRF optimization step that refines the NeRF scene model and those camera poses. Consequently, our learning algorithm also iterates over the NeRF estimation and optimization step to further overcome local minima between the two phases (AB...AB). 
 
In the following, we first present our pose-free NeRF estimation procedure in Sec~\ref{sec:gan}, and then introduce the regularized and iterative NeRF optimization step in Sec~\ref{sec:iterative_step}. The training strategy is detailed in Sec~\ref{sec:training} and model architecture is detailed in Sec~\ref{sec:implementation}.

\subsection{Pose-free NeRF Estimation}\label{sec:gan}
As the initial stage of our method, in phase A, we do not have a reasonable camera pose estimation for each image or a pre-trained radiance field. Our goal for this stage is to predict a rough pose for each image and also learn a rough radiance field of the scene. As shown in the left part of Fig.~\ref{fig:pipeline}, we use adversarial learning to achieve the goals. Our architecture contains two parts: a generator $G$ and a discriminator $D$. Taking a random camera pose $\phi$ as input, the generator $G$ will synthesize the image observed at the view by querying the neural radiance field and performing NeRF-like volume rendering. The set of synthesized images from many sampled camera poses will be decomposed into patches and compared against the set of real patches by the discriminator $D$. The fake and real patches are sampled via the dynamic patch sampling strategy which will be described in Sec~\ref{sec:training}. $G$ and $D$ are trained adversarially, as is done by the classical GAN work~\cite{goodfellow2014generative}. This adversarial training allows us to roughly learn the radiance field and estimate camera poses at random initialization.

Formally, we minimize a distribution distance between the real image patches $P_d(I)$ from the training set $\mathcal{I}$ and the generated image patches $P_g(I|\Theta)$, which are defined as below:
\begin{align}
\Theta^* &= \arg\min_{\Theta}{Dist}\left(P_g(I|\Theta)||P_d(I)\right)\\
P_g(I|\Theta) &= \int_{\phi} G(\phi;F_\Theta) P(\phi) d\phi
\end{align}

To minimize the distribution distance, we adopt the following GAN learning framework based on an adversarial loss $\mathcal{L}_A$ defined as follows: 
\begin{align}
\min_\Theta\max_\eta \mathcal{L}_A(\Theta,\eta) = &\mathbb{E}_{I\sim P_d}[\log(D(I;\eta))]\nonumber \\+&\mathbb{E}_{\hat{I}\sim P_g}[\log(1-D(\hat{I};\eta))]
\end{align}

where $\eta$ are the network parameters of the discriminator $D$ and $\mathbb{E}$ denotes expectation. 

Along with the two standard components, we train an inversion network $E$ that maps image patches to the corresponding camera poses. We train the inversion network with the pairs of randomly sampled camera poses and generated image patches. The image patches are deterministically sampled from original images via a static sampling strategy which will be described in Sec~\ref{sec:training}. The inputs of the inversion network are these image patches, and the outputs are the corresponding camera poses. Formally, we denote the parameters of the inversion network $E$ as $\theta_{E}$, and its loss function can be written as,
\begin{align}
\mathcal{L}_{E}(\theta_{E}) = \mathbb{E}_{\phi\sim P(\phi)} \left[\|E(G(\phi; F_\Theta);\theta_E) - \phi\|^2_2\right]
\end{align}
We note that the inversion network is trained in a self-supervised manner, which exploits the synthetic image patches and their corresponding camera poses as the training data. With the increasingly better-trained generator, the inversion network would be able to predict camera poses for real image patches. After the overall training is converged, we apply the inverse network to generate camera pose estimates $\{\phi_i^{'}=E(I_i), I_i\in \mathcal{I}\}$ for the training set $\mathcal{I}$.

\subsection{Regularized Learning Strategy}\label{sec:iterative_step} 
After the pose-free NeRF estimation step, we obtain an initial NeRF model and camera pose estimates for the training images. Due to the sparse sampling of the input image patches and the constrained capability of the inversion network, neither the NeRF representation nor the estimated camera poses $\Phi'=\{\phi_i^{'}\}$ are accurate enough. However, they provide a good initialization for the overall training procedure. This allows us to introduce a refinement step for the NeRF model and camera poses, phase B, as illustrated in the right part of Fig.~\ref{fig:pipeline}. Specifically, this phase optimizes the pose embedding and the NeRF model by minimizing the photometric reconstruction error $\mathcal{L}_N(\Theta,\Phi)$ as defined in Eqn.~\ref{nerf}. 

We note that existing work like iNeRF and NeRF$--$ can search a limited scope in the pose space during NeRF optimization. However, the pose optimization problem in the standard NeRF model is highly non-convex, and hence their results strongly depend on camera pose initialization and are still insufficient for our challenging test scenarios. To mitigate this issue, we propose a regularized learning strategy (AB $\dots$AB) by interleaving the pose-free NeRF estimation step (phase A) and the NeRF refinement step (phase B) to further improve the quality of the NeRF model and pose estimation. Such a design is based on our empirical findings that the pose-free NeRF estimation can also improve NeRF model and camera poses from the refinement step. 

This strategy regularizes the gradient descent-based model optimization by the pose prediction from the learned inversion network. Intuitively, with the adversarial training of the NeRF model, the domain gap between synthesized fake images and true images is narrowing, so those pose predictions provide a reasonable and effective constraint for the joint radiance fields and pose optimization. Formally, we define a hybrid loss function $\mathcal{L}_R$ that combines the photometric reconstruction errors and an L2 loss penalizing the deviation from the predictions of the inversion network, which can be written as below,
\begin{align}
\mathcal{L}_R(\Theta,\Phi)= \mathcal{L}_N(\Theta,\Phi) + \frac{\lambda}{n}\sum_{i=1}^n \| E(I_i; \theta_E) - \phi_i \|^2_2 \label{hybrid_optimization}
\end{align}
where $\lambda$ is the weighting coefficient and $\mathcal{L}_N(\Theta,\Phi)$ is the photometric loss defined in Eqn.~\ref{nerf}.

\begin{table*}[t]
	\centering
	\begin{tabular}{*{14}{c}}
		\toprule
		\multirow{2}{*}{Data} & \multirow{2}{*}{Scene} & \multicolumn{4}{c}{$\uparrow$ PSNR}  & \multicolumn{4}{c}{$\uparrow$ SSIM} & \multicolumn{4}{c}{$\downarrow$ LPIPS} \\
		
		\cmidrule(lr){3-6}\cmidrule(lr){7-10}\cmidrule(lr){11-14} 
		
		&& \multicolumn{1}{c}{\textbf{C+n}} & \multicolumn{1}{c}{\textbf{C+r}} & \multicolumn{1}{c}{Ours} & \multicolumn{1}{c}{\textbf{G+n}}
		& \multicolumn{1}{c}{\textbf{C+n}} & \multicolumn{1}{c}{\textbf{C+r}} & \multicolumn{1}{c}{Ours} & \multicolumn{1}{c}{\textbf{G+n}}
		& \multicolumn{1}{c}{\textbf{C+n}} & \multicolumn{1}{c}{\textbf{C+r}} & \multicolumn{1}{c}{Ours} & \multicolumn{1}{c}{\textbf{G+n}} \\
		
		\midrule
		\multirow{6}{*}{\rotatebox[origin=c]{90}{\parbox[c]{1cm}{\centering Synthetic-NeRF}}} & Chair & 33.75&32.70&31.30&32.84    &0.97&0.95&0.94&0.97    &0.03&0.05&0.08&0.04 \\
		& Drums &22.39&23.42&24.30&26.71    &0.91&0.88&0.90&0.93    &0.10&0.13&0.13&0.07 \\
		& Hotdog &25.14&33.59&32.00&29.72   &0.96&0.97&0.96&0.95    &0.05&0.03&0.07&0.04 \\
		& Lego &29.13&28.73&28.52&31.06     &0.93&0.92&0.91&0.95    &0.06&0.08&0.09&0.04 \\
		& Mic &26.62&31.58&31.07&34.65      &0.96&0.97&0.96&0.97    &0.04&0.03&0.06&0.02 \\
		& Ship &27.49&28.04&26.51&28.97     &0.88&0.86&0.85&0.82    &0.16&0.18&0.21&0.15 \\
		\midrule
		\multirow{4}{*}{\rotatebox[origin=c]{90}{\parbox[c]{1cm}{\centering DTU}}}  

		& Scan4 &22.05&24.23&22.88&25.52    &0.69&0.72&0.82&0.78   &0.32&0.20&0.37&0.18 \\
		& Scan48 &6.718&10.40&23.25&26.20     &0.52&0.62&0.87&0.90    &0.65&0.60&0.21&0.21 \\
		& Scan63 &27.80&26.61&25.11&32.19     &0.90&0.90&0.90&0.93    &0.21&0.19&0.29&0.24 \\

		& Scan104 &10.52&13.92&21.40&23.35    &0.48&0.55&0.76&0.82  &0.60&0.59&0.44&0.36 \\
		\bottomrule
	\end{tabular}
	\caption{\textbf{Quantitative comparison among COLMAP-based NeRF~\cite{nerf} (C+n), COLMAP-based NeRF with additional refinement (C+r), NeRF with ground-truth poses(G+n), and ours on the Synthetic-NeRF~\cite{nerf} dataset and DTU~\cite{jensen2014large} dataset. } We report PSNR, SSIM and LPIPS metrics to evaluate novel view synthesis quality. Our method without posed camera generates novel views on par with COLMAP-based NeRF and is more robust to challenging scene where COLMAP-based NeRF fails. }
	\label{compare_1}
\end{table*}

\begin{figure*}[t]
	\begin{center}
		\includegraphics[width=1.0\linewidth]{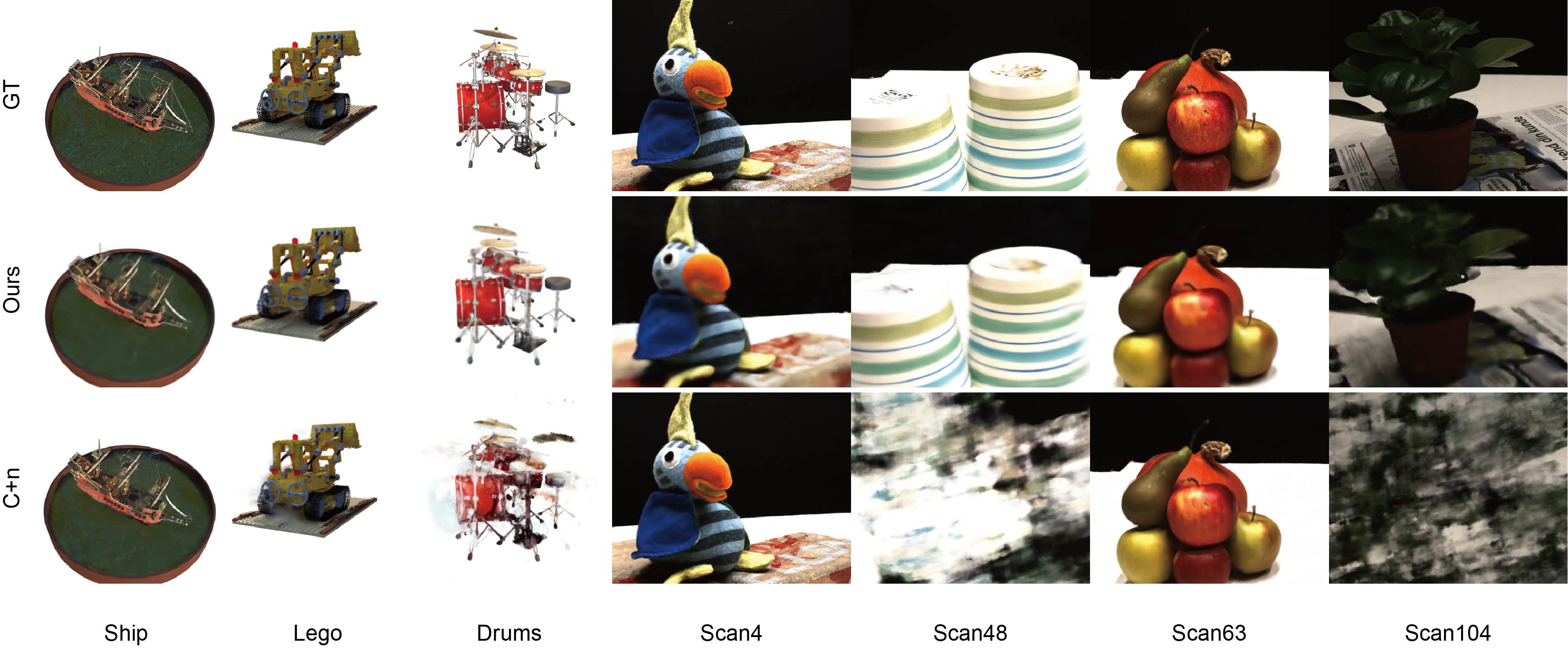}
	\end{center}
  	\vspace{-0.1in}
	\caption{\textbf{Qualitative comparison between COLMAP-based NeRF (C+n) and ours on novel view synthesis quality on Synthetic-NeRF~\cite{nerf} dataset and DTU~\cite{jensen2014large} dataset. } `GT' means ground-truth images.}
	\label{fig:campare_vis_all}
\end{figure*} 

\subsection{Training}\label{sec:training}
Initially, we set all camera extrinsics to be an identity matrix. In phase A, we sample camera poses $\phi$ randomly from the prior pose distribution. In the Synthetic-NeRF dataset, the cameras are uniformly distributed at the upper hemisphere and towards the origin. In practice, we compute the rotation matrix directly from the camera position and the lookat point. In the DTU dataset, the cameras are uniformly distributed at the upper hemisphere with an azimuth range of $[0, 150]$, and the lookat point is distributed at a gaussian distribution $\mathcal{N}(0, 0.01^2)$. We analyze how the mismatch of prior pose distribution influences the performance in the supplement material.	

To train the generative radiance field, we follow a similar patch sampling strategy as GRAF~\cite{schwarz2020graf} for computation and memory efficiency. Specifically, for the GAN training process, we adopt a dynamic patch sampling strategy, as is illustrated in the lower left part of Fig.~\ref{fig:pipeline}. Each patch is sampled within the image domain with a fixed size of $16 \times 16$ but dynamic scale and random offset. For the pose optimization process, we adopt a static patch sampling strategy, as is illustrated in the upper left part of Fig.~\ref{fig:pipeline}. Each patch is uniformly sampled across the whole image domain with a fixed size of $64 \times 64$. This sampling strategy uniquely represents the whole image with a sparse patch with which we estimate the corresponding camera pose. We also scale the camera intrinsics at the beginning to maximize the receptive field and progressively increase it to the original value to concentrate on fine details. In practice, these strategies bring great benefits to the stability of the GAN training process.

\subsection{Implementation Details}\label{sec:implementation} 
We adopt the network architecture of the original NeRF~\cite{nerf} and its hierarchical sampling strategy to our generator. The numbers of sampled points of both coarse sampling and importance sampling are set to $64$. Differently, because the GAN training only narrows the distribution of real patches and fake patches (``coarse'' and ``fine''), we utilize the same MLPs in hierarchical sampling strategy to ensure the pose spaces of ``coarse'' and ``fine'' networks are aligned. For a fair comparison, we increase the dimension of the MLPs from the original 256 to 360 to keep the overall parameters unchanged. The discriminator network follows GRAF~\cite{schwarz2020graf}, in which instance normalization~\cite{ulyanov2016instance} over features and spectral normalization~\cite{miyato2018spectral} over weights are applied. We borrow the Vision Transformer Network~\cite{dosovitskiy2020image} to build our inversion network, whose last layer is modified to output a camera pose.

We use RMSprop~\cite{kingma2013auto} algorithm to optimize the generator and the discriminator with learning rates of 0.0005 and 0.0001, respectively. As for the inversion network and camera poses, we use Adam~\cite{kingma2014adam} algorithm with learning rates of 0.0001 and 0.005.

\section{Experiments}

Here we compare our method with other approaches which require camera poses or a coarse camera initialization on view synthesis task and evaluate our method in various scenarios.
We run our experiments on a PC with Intel i7-8700K CPU, $32$GB RAM, and a single Nvidia RTX TITAN GPU, where our approach takes 30 hours to train the network on a single scene. 

\subsection{Performance Evaluations}
\paragraph{Novel View Synthesis Comparison} We firstly compare novel view synthesis quality on the Synthetic-NeRF~\cite{nerf} and DTU~\cite{jensen2014large} datasets with three other approaches: Original NeRF~\cite{nerf} with precalibrated camera poses from COLMAP~\cite{schonberger2016structure}, denoted by \textbf{C+n}; Original NeRF with precalibrated camera poses from COLMAP but jointly refined via gradient descent, denoted by \textbf{C+r}; Original NeRF with ground-truth camera poses, denoted by \textbf{G+n}. We report the standard image quality metrics Peak Signal-to-Noise Ratio (PSNR), SSIM~\cite{wang2004image} and LPIPS~\cite{zhang2018unreasonable} to evaluate image perceptual quality.

For evaluation, we need to estimate the camera poses of the test view images. Since our method can predict poses of new images, the camera poses of test view are directly estimated by our well-trained model. Conversely, for the COLMAP-based methods, we need to estimate the camera poses of images in the training set and test set together to keep them lie in the same space. We note that the COLMAP produces more accurate poses estimation with more input images, so for fair evaluation, we only choose a limited number of test images. The selection is based on maximizing their mutual angular distance between views so that test samples can cover different perspectives of the object as much as possible. For the Synthetic-NeRF dataset, we follow the same split as the original but randomly sample eight images from test set for testing. The COLMAP is incapable to register the images with the resolution of $[400, 400]$ as shown in the supplement material, so we provide COLMAP with 108 images of $800\times800$ for camera registration which COLMAP performs much better. The training image resolution for all the methods is $400 \times 400$. For the DTU dataset, we use four representative scenes, on each of which we take every 8-th image as test images and take the rest 43 images for training. The input image resolution is $500 \times 400$. The scene selection is based on consideration of diversity: synthetic scenes (Synthetic-NeRF); real scenes with rich textures (scan4 and scan63); real scenes with less texture (scan48 and scan104).

As in Fig.~\ref{fig:campare_vis_all}, we show{\tiny } the visualization comparison with methods on the Synthetic-NeRF and DTU datasets. Our method outperforms the \textbf{C+n} in challenging scenes, e.g., scan48, scan104, lego, and drums while achieving similar results on regular scenes with enough keypoints. These challenging scenes do not have enough keypoints for pose estimation, so make NeRF which needs precise poses as input fail to synthesis good results. 

We also show the quantitative performance of all the three methods in Tab.~\ref{compare_1} on the Synthetic-NeRF and DTU datasets. We notice that our method outperforms the \textbf{C+n} in those challenging scenes. For other scenes, our method generates satisfactory results on par with the COLMAP-based NeRF methods. \textbf{C+r} have a better performance than \textbf{C+n}'s. However, limited by the poor pose initialization, \textbf{C+r} can not produce the same performance as ours in some challenging scenes. 

\begin{table}[t]
	\centering
	\begin{tabular}{*{4}{c}}
		\toprule
		Methods & Scan48 & Scan97 & Scan104 \\
		\midrule
		IDR(masked)~\cite{yariv2020multiview}             & 21.17 & 17.42 & 12.26 \\
		Ours(masked)    & 20.40 & 19.40 & 19.81 \\		
		Ours            & 25.71 & 24.52 & 25.70 \\
		\bottomrule
	\end{tabular}
	\caption{\textbf{Quantitiative rendering quality comparison between IDR and ours on DTU~\cite{jensen2014large} dataset. }The evaluation metric is PSNR.}
	\label{compare_2}
\end{table}

\begin{figure}[t]
	\begin{center}
		\includegraphics[width=1.0\linewidth]{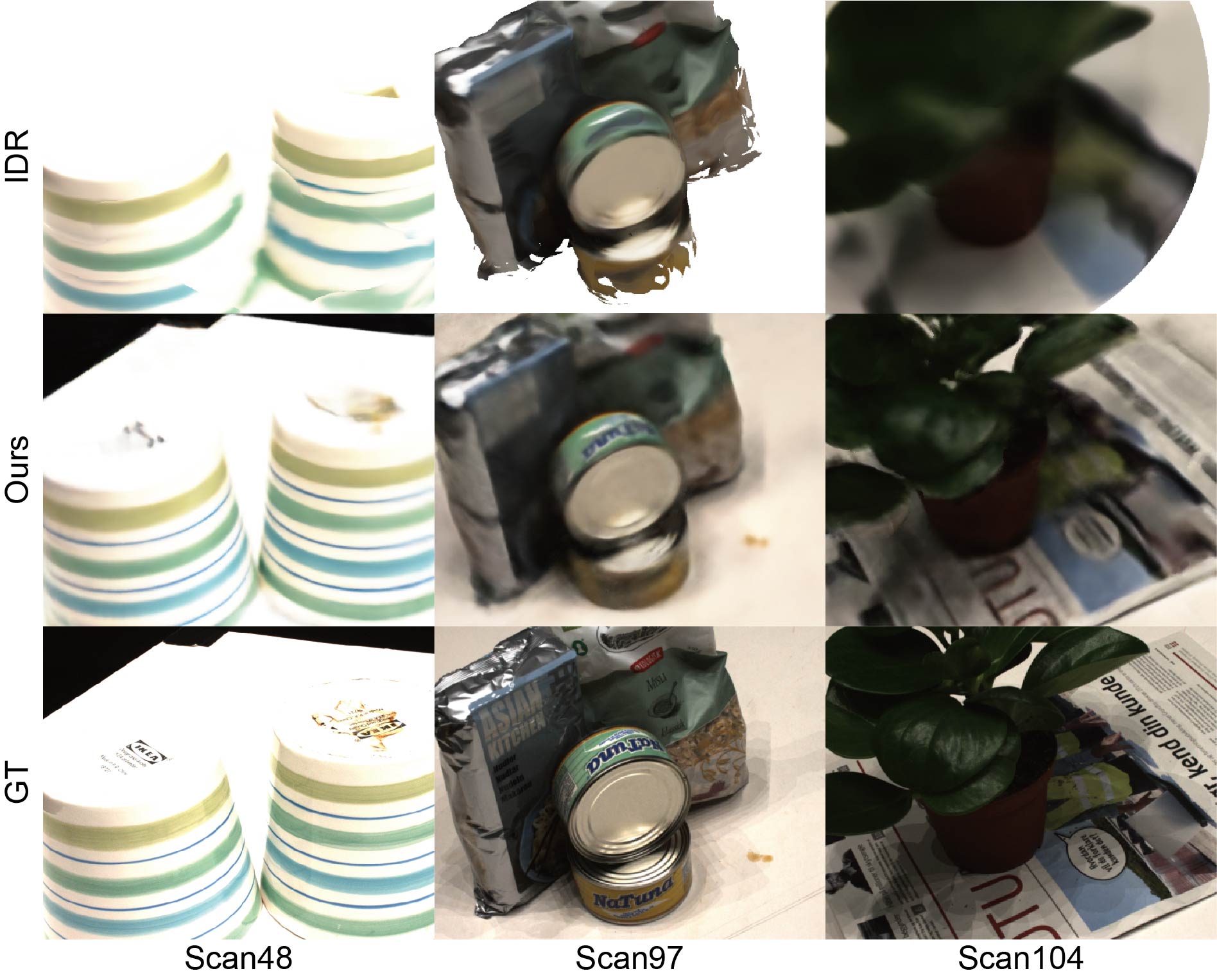}
	\end{center}
	\vspace{-0.1in}
	\caption{\textbf{Qualitative rendering quality comparison between IDR~\cite{yariv2020multiview} and ours on DTU dataset. }}
	\label{fig:compare_vis_idr}
\end{figure}

Additionally, to further demonstrate our architecture's ability to learn the high-quality 3D representation without camera poses, we also compare with the state-of-the-art 3D surface reconstruction method, IDR~\cite{yariv2020multiview}, by comparing the rendering quality. Note that the IDR method requires image masks and noisy camera initializations, while our method does not need them. We follow the same setting of optimizing the model and camera extrinsics jointly on 49 training images of each scene and report the mean PSNR as evaluation metrics. We report the PSNR computed on the whole image and within the mask, which is the same evaluation protocol as IDR. The qualitative and quantitative results are in Tab.~\ref{compare_2} and Fig.~\ref{fig:compare_vis_idr}. It can be seen that our volume-rendering-based method produces more natural images, while IDR produces results with more artifacts and fewer fine details. 

\vspace{-3mm}
\paragraph{Camera Poses Comparison} We evaluate the accuracy of camera poses estimation on the Synthetic-NeRF dataset which contains several relatively challenging scenes with repeated patterns or less texture. The camera model of COLMAP is SIMPLE PINHOLE with shared intrinsics, $f = 1111.111$, $cx = 400$, $cy = 400$. For COLMAP, the input image size is $800\times800$ and the number is $108$, while for our method, the input image size is $400\times400$ and the number is $100$. We note that COLMAP produces more accurate estimates with more input images. In Tab.~\ref{cam_pose_tab}, we report the mean translation and rotation difference on the train set computed with the ATE toolbox~\cite{zhang2018tutorial}. Our method outperforms the COLMAP~\cite{schonberger2016structure} on the drums and lego scenes which have less texture and repeated patterns. However, on the other scenes, which still contain enough reliable keypoints, our method is not accurate as the COLMAP. 

\subsection{Ablation Study}
\begin{table}[t]
	\centering
	\begin{tabular}{*{5}{c}}
		\toprule
		\multirow{2}{*}{Scene} & \multicolumn{2}{c}{COLMAP~\cite{schonberger2016structure}} & \multicolumn{2}{c}{Ours} \\
		
		\cmidrule(lr){2-3}\cmidrule(lr){4-5}
		
		& \multicolumn{1}{c}{$\downarrow$ Rot(deg)} & \multicolumn{1}{c}{$\downarrow$ Trans} & \multicolumn{1}{c}{$\downarrow$ Rot(deg)} & \multicolumn{1}{c}{$\downarrow$ Trans} \\
		
		\midrule
		 Chair & 0.119 & 0.006  & 0.363 & 0.018 \\
		Drums & 9.985 & 0.522   & 0.204 & 0.010 \\
		Hotdog & 0.542 & 0.024  & 2.349 & 0.122 \\
		Lego & 7.492 & 0.332   & 0.430 & 0.023 \\
	    Mic & 0.746  & 0.047    & 1.865 & 0.031 \\
		Ship & 0.191 & 0.010    & 3.721 & 0.176 \\
		\bottomrule
	\end{tabular}
	\caption{\textbf{Quantitative camera poses accuracy comparison between COLMAP and ours on Synthetic-NeRF~\cite{nerf} dataset.} We report the mean camera rotation difference (Rot) and translation difference (Trans) over the training set. }
	\label{cam_pose_tab}
\end{table}

\begin{figure}[t]
	\begin{center}
		\includegraphics[width=1.0\linewidth]{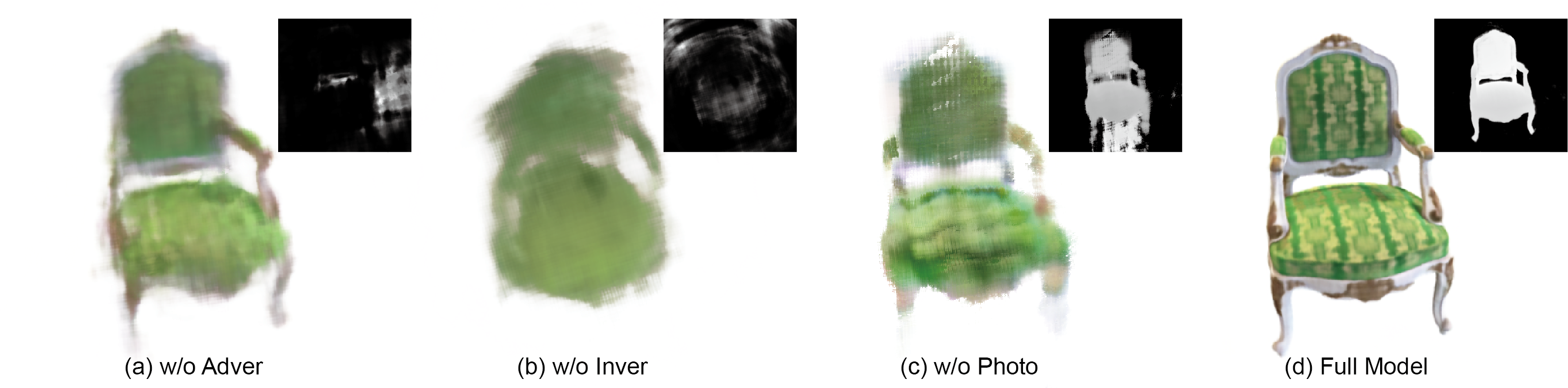}
	\end{center}
	\vspace{-0.1in}
	\caption{\textbf{Ablation study.} We visualize novel view RGB images and depth maps of the four different configurations. }
	\label{fig:ablation_fig_1}
\end{figure}

In Tab.~\ref{ablation_tab_1} and Fig.~\ref{fig:ablation_fig_1}, we show an ablation study over different components of our model. Our full architecture of the combination of adversarial training, inversion network, and photometric loss achieves the best performance. Without either the adversarial loss or the inversion network, the model is incapable to learn correct geometry; without the photometric loss, the model is only capable to get coarse radiance fields. 

\begin{table}[t]
	\centering
	\begin{tabular}{cccccc}
		\toprule
		\multicolumn{1}{c}{Adver} & 
		\multicolumn{1}{c}{Inver} &
		\multicolumn{1}{c}{Photo} &
		\multicolumn{1}{c}{$\uparrow$ PSNR} &
		\multicolumn{1}{c}{$\downarrow$ Rot(deg)} &
		\multicolumn{1}{c}{$\downarrow$ Trans}
		 \\
		\midrule
  	     & \checkmark  &  \checkmark  &  19.31 & 108.22 & 2.53 \\
  	    \checkmark &   &  \checkmark  &  13.82 & 132.85 & 3.05 \\
        \checkmark & \checkmark  &    & 20.60 & 5.91 & 0.24 \\
		\checkmark & \checkmark & \checkmark & 31.30 & 0.36 & 0.02 \\
		\bottomrule
	\end{tabular}
	\caption{\textbf{Ablation study.} We report PSNR, camera rotation difference (Rot), and translation difference (Trans) of the full model (the last row) and three configurations by removing the adversarial loss (Adver), the inversion network (Inver), and the photometric loss (Photo), respectively. Removing adversarial loss and inversion network prevents the model from learning reasonable camera poses. Removing photometric loss prevents the model from getting accurate camera poses.}
	\label{ablation_tab_1}
\end{table}

\begin{table}[t]
	\centering
	\begin{tabular}{ccccc}
		\toprule
		\multicolumn{1}{c}{A, B} & 
		\multicolumn{1}{c}{A, AB...AB, B} &
		\multicolumn{1}{c}{$\uparrow$ PSNR} &
		\multicolumn{1}{c}{$\downarrow$ Rot(deg)} &
		\multicolumn{1}{c}{$\downarrow$ Trans}
		 \\
		\midrule
  	    \checkmark  & & 29.23  &  0.592  & 0.034 \\{\tiny }
		 & \checkmark & 31.30 & 0.363 & 0.018 \\
		\bottomrule
	\end{tabular}
	\caption{\textbf{Optimization schemes analysis.} We compare two optimization schemes: `A, B' and `A, AB...AB, B'. The additional iterative optimization step enables our model to achieve much better results.}.
	\label{ablation_tab_2}
\end{table}

\begin{figure}[t]
	\begin{center}
		\includegraphics[width=1.0\linewidth]{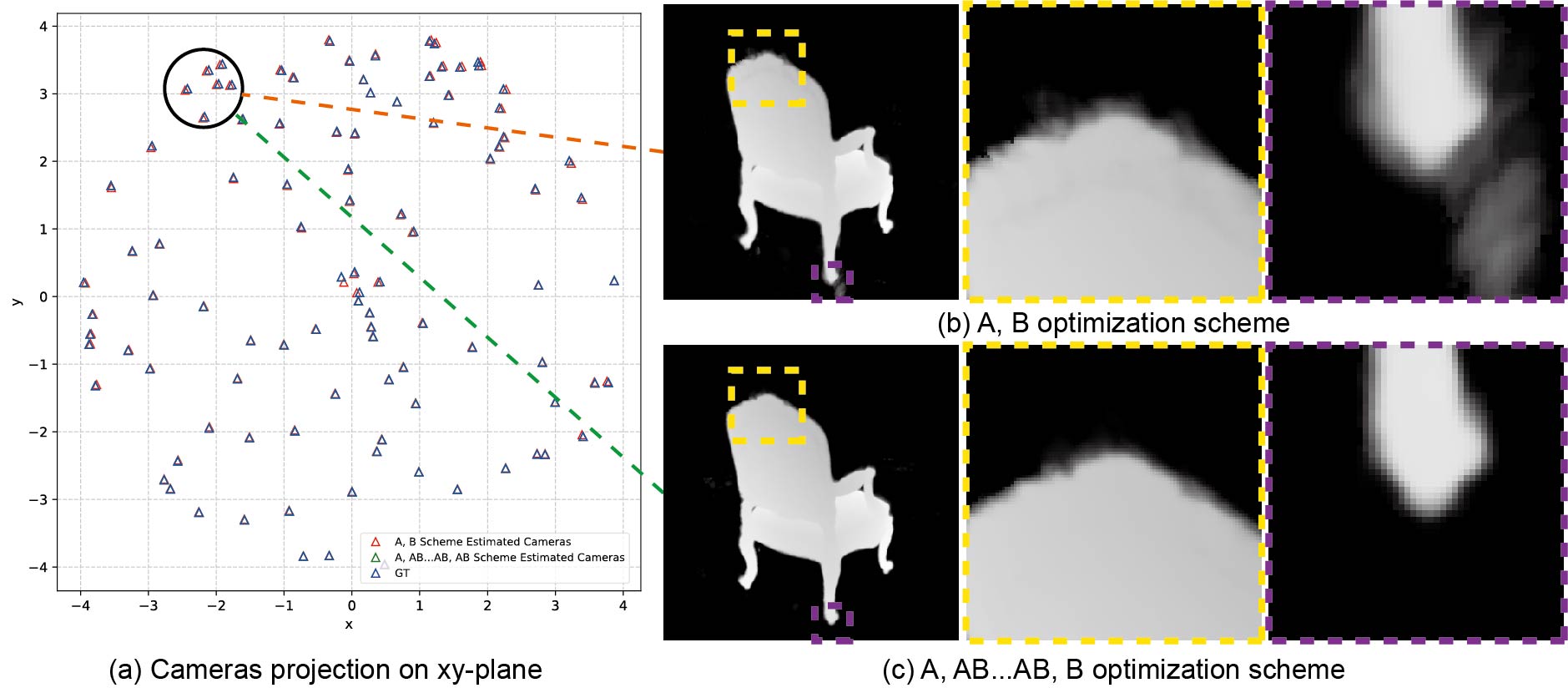}
	\end{center}
	\vspace{-0.1in}
	\caption{\textbf{Optimization schemes analysis.} On the left, we visualize the projection of camera poses on $xy$-plane of the obtained image from the two optimization schemes. On the right, we show depth maps of the view in the circled camera region and two detailed parts (yellow and purple insets) of them. }
	\label{fig:ablation_tab_2}
\end{figure}

In Tab.~\ref{ablation_tab_2} and Fig.~\ref{fig:ablation_tab_2}, we analyze different optimization schemes. We represent Phase A and Phase B as A and B respectively. Our adopted iterative optimization scheme on the pattern `A, AB...AB, B' achieves much higher image quality and camera pose accuracy than that of `A, B'. In Fig.~\ref{fig:ablation_tab_2}, the iterative optimization scheme gets much finer geometry along the edge, and the estimated camera poses align much closer to the ground-truth camera poses. These results demonstrate that the iterative learning strategy can further help overcome local minima.

\section{Discussion and Conclusion}
\myparagraph{Discussion} First, our method does not depend on camera pose initialization, but it does require a reasonable camera pose sampling distribution. For different datasets, we rely on a camera sampling distribution not far from the true distribution to alleviate the difficulties for radiance field estimation. This could potentially be mitigated by learning the underlying pose sampling space automatically. A promising future direction would be combining global appearance distribution optimization (our approach) and local feature matching (pose distribution estimator) for the appearance and geometric reconstruction in an end-to-end manner. This combination potentially preserves our capability to challenging cases and relax to more general scenes without accurate distribution prior.
Secondly, jointly optimizing camera poses and scene representation is a challenging task and opt to fall in local minima. Although in real datasets, we achieve good novel view synthesis quality on par with NeRF if the accurate camera poses present, our optimized camera poses are still not so accurate as of the COLMAP when there are sufficient amount of reliable keypoints. This may be due to that our inversion network, which maps images to camera poses, could only take in image patches with limited size for computation efficiency. This might be fixed by importance sampling. 

\myparagraph{Conclusion} We have presented GNeRF, a GAN-based framework to reconstruct neural radiance fields and estimate camera poses when the camera poses are completely unknown and scene conditions can be complicated.
Our framework is fully differentiable and end-to-end trainable.
Specifically, our first phase enables GAN-based joint optimization for the 3D representation and the camera poses, and our hybrid and iterative scheme by interleaving the first and second phases would further refine the results robustly.
Extensive experiments demonstrate the effectiveness of our approach.
Impressively, our approach has demonstrated promising results on those scenes with repeated patterns or even low textures, which have been regarded as extremely challenging before. 
We believe our approach is a critical step towards the more general neural scene modeling goal using less human-crafted priors.

\section*{Acknowledgements}
We would like to thank the anonymous reviewers for their detailed and constructive comments which were helpful in refining the paper. This work was supported by NSFC programs (61976138, 61977047), the National Key Research and Development Program (2018YFB2100500), STCSM (2015F0203-000-06) and SHMEC (2019-01-07-00-01-E00003)
	
{\small
	\bibliographystyle{ieee_fullname}
	\bibliography{reference}

\begin{thebibliography}{10}\itemsep=-1pt

\bibitem{andrew2001multiple}
Alex~M Andrew.
\newblock Multiple view geometry in computer vision.
\newblock {\em Kybernetes}, 2001.

\bibitem{balntas2018relocnet}
Vassileios Balntas, Shuda Li, and Victor Prisacariu.
\newblock Relocnet: Continuous metric learning relocalisation using neural
  nets.
\newblock In {\em Proceedings of the European Conference on Computer Vision
  (ECCV)}, 2018.

\bibitem{brahmbhatt2018geometry}
Samarth Brahmbhatt, Jinwei Gu, Kihwan Kim, James Hays, and Jan Kautz.
\newblock Geometry-aware learning of maps for camera localization.
\newblock In {\em Proceedings of the IEEE Conference on Computer Vision and
  Pattern Recognition (CVPR)}, 2018.

\bibitem{chan2020pi}
Eric~R Chan, Marco Monteiro, Petr Kellnhofer, Jiajun Wu, and Gordon Wetzstein.
\newblock pi-gan: Periodic implicit generative adversarial networks for
  3d-aware image synthesis.
\newblock In {\em Proceedings of the IEEE Conference on Computer Vision and
  Pattern Recognition (CVPR)}, 2021.

\bibitem{chen2020free}
Anpei Chen, Ruiyang Liu, Ling Xie, and Jingyi Yu.
\newblock A free viewpoint portrait generator with dynamic styling.
\newblock {\em ACM Trans. Graph.}, 2021.

\bibitem{mvsnerf}
Anpei Chen, Zexiang Xu, Fuqiang Zhao, Xiaoshuai Zhang, Fanbo Xiang, Jingyi Yu,
  and Hao Su.
\newblock Mvsnerf: Fast generalizable radiance field reconstruction from
  multi-view stereo.
\newblock {\em arXiv preprint arXiv:2103.15595}, 2021.

\bibitem{choy20163d}
Christopher~B Choy, Danfei Xu, JunYoung Gwak, Kevin Chen, and Silvio Savarese.
\newblock 3d-r2n2: A unified approach for single and multi-view 3d object
  reconstruction.
\newblock In {\em Proceedings of the European Conference on Computer Vision
  (ECCV)}, 2016.

\bibitem{chum2007total}
Ondrej Chum, James Philbin, Josef Sivic, Michael Isard, and Andrew Zisserman.
\newblock Total recall: Automatic query expansion with a generative feature
  model for object retrieval.
\newblock In {\em The IEEE International Conference on Computer Vision (ICCV)},
  2007.

\bibitem{dosovitskiy2020image}
Alexey Dosovitskiy, Lucas Beyer, Alexander Kolesnikov, Dirk Weissenborn,
  Xiaohua Zhai, Thomas Unterthiner, Mostafa Dehghani, Matthias Minderer, Georg
  Heigold, Sylvain Gelly, et~al.
\newblock An image is worth 16x16 words: Transformers for image recognition at
  scale.
\newblock {\em arXiv preprint arXiv:2010.11929}, 2020.

\bibitem{fan2017point}
Haoqiang Fan, Hao Su, and Leonidas~J Guibas.
\newblock A point set generation network for 3d object reconstruction from a
  single image.
\newblock In {\em Proceedings of the IEEE Conference on Computer Vision and
  Pattern Recognition (CVPR)}, 2017.

\bibitem{faugeras2001geometry}
Olivier Faugeras, Quang-Tuan Luong, and Theo Papadopoulo.
\newblock {\em The geometry of multiple images: the laws that govern the
  formation of multiple images of a scene and some of their applications}.
\newblock MIT press, 2001.

\bibitem{fischler1981random}
Martin~A Fischler and Robert~C Bolles.
\newblock Random sample consensus: a paradigm for model fitting with
  applications to image analysis and automated cartography.
\newblock {\em Communications of the ACM}, 1981.

\bibitem{gkioxari2019mesh}
Georgia Gkioxari, Jitendra Malik, and Justin Johnson.
\newblock Mesh r-cnn.
\newblock In {\em The IEEE International Conference on Computer Vision (ICCV)},
  2019.

\bibitem{goodfellow2014generative}
Ian Goodfellow, Jean Pouget-Abadie, Mehdi Mirza, Bing Xu, David Warde-Farley,
  Sherjil Ozair, Aaron Courville, and Yoshua Bengio.
\newblock Generative adversarial nets.
\newblock {\em Neural Information Processing Systems (NeurIPS)}, 2014.

\bibitem{henzler2019escaping}
Philipp Henzler, Niloy~J Mitra, and Tobias Ritschel.
\newblock Escaping plato's cave: 3d shape from adversarial rendering.
\newblock In {\em The IEEE International Conference on Computer Vision (ICCV)},
  2019.

\bibitem{irschara2009structure}
Arnold Irschara, Christopher Zach, Jan-Michael Frahm, and Horst Bischof.
\newblock From structure-from-motion point clouds to fast location recognition.
\newblock In {\em Proceedings of the IEEE Conference on Computer Vision and
  Pattern Recognition (CVPR)}, 2009.

\bibitem{jensen2014large}
Rasmus Jensen, Anders Dahl, George Vogiatzis, Engin Tola, and Henrik Aan{\ae}s.
\newblock Large scale multi-view stereopsis evaluation.
\newblock In {\em Proceedings of the IEEE Conference on Computer Vision and
  Pattern Recognition (CVPR)}, 2014.

\bibitem{zhang2021stnerf}
Zhang Jiakai, Liu Xinhang, Ye Xinyi, Zhao Fuqiang, Zhang Yanshun, Wu Minye,
  Zhang Yingliang, Xu Lan, and Yu Jingyi.
\newblock Editable free-viewpoint video using a layered neural representation.
\newblock In {\em ACM SIGGRAPH computer graphics}, 2021.

\bibitem{kajiya1984ray}
James~T Kajiya and Brian~P Von~Herzen.
\newblock Ray tracing volume densities.
\newblock {\em ACM SIGGRAPH computer graphics}, 18(3):165--174, 1984.

\bibitem{karras2019style}
Tero Karras, Samuli Laine, and Timo Aila.
\newblock A style-based generator architecture for generative adversarial
  networks.
\newblock In {\em Proceedings of the IEEE Conference on Computer Vision and
  Pattern Recognition (CVPR)}, 2019.

\bibitem{kendall2017geometric}
Alex Kendall and Roberto Cipolla.
\newblock Geometric loss functions for camera pose regression with deep
  learning.
\newblock In {\em Proceedings of the IEEE Conference on Computer Vision and
  Pattern Recognition (CVPR)}, 2017.

\bibitem{kendall2015posenet}
Alex Kendall, Matthew Grimes, and Roberto Cipolla.
\newblock Posenet: A convolutional network for real-time 6-dof camera
  relocalization.
\newblock In {\em The IEEE International Conference on Computer Vision (ICCV)},
  2015.

\bibitem{kingma2014adam}
Diederik~P Kingma and Jimmy Ba.
\newblock Adam: A method for stochastic optimization.
\newblock In {\em International Conference on Learning Representations (ICLR)},
  2015.

\bibitem{kingma2013auto}
Diederik~P Kingma and Max Welling.
\newblock Auto-encoding variational bayes.
\newblock In {\em International Conference on Learning Representations (ICLR)},
  2013.

\bibitem{laskar2017camera}
Zakaria Laskar, Iaroslav Melekhov, Surya Kalia, and Juho Kannala.
\newblock Camera relocalization by computing pairwise relative poses using
  convolutional neural network.
\newblock In {\em Proceedings of the IEEE International Conference on Computer
  Vision Workshops}, 2017.

\bibitem{liu2020neural}
Lingjie Liu, Jiatao Gu, Kyaw~Zaw Lin, Tat-Seng Chua, and Christian Theobalt.
\newblock Neural sparse voxel fields.
\newblock In {\em Neural Information Processing Systems (NeurIPS)}, 2020.

\bibitem{liu2019learning}
Shichen Liu, Shunsuke Saito, Weikai Chen, and Hao Li.
\newblock Learning to infer implicit surfaces without 3d supervision.
\newblock In {\em Neural Information Processing Systems (NeurIPS)}, 2019.

\bibitem{lorensen1987marching}
William~E Lorensen and Harvey~E Cline.
\newblock Marching cubes: A high resolution 3d surface construction algorithm.
\newblock In {\em ACM siggraph computer graphics}, 1987.

\bibitem{lowe2004distinctive}
David~G Lowe.
\newblock Distinctive image features from scale-invariant keypoints.
\newblock {\em International journal of computer vision}, 60(2):91--110, 2004.

\bibitem{martin2020nerf}
Ricardo Martin-Brualla, Noha Radwan, Mehdi~SM Sajjadi, Jonathan~T Barron,
  Alexey Dosovitskiy, and Daniel Duckworth.
\newblock Nerf in the wild: Neural radiance fields for unconstrained photo
  collections.
\newblock In {\em Proceedings of the IEEE Conference on Computer Vision and
  Pattern Recognition (CVPR)}, 2021.

\bibitem{mescheder2019occupancy}
Lars Mescheder, Michael Oechsle, Michael Niemeyer, Sebastian Nowozin, and
  Andreas Geiger.
\newblock Occupancy networks: Learning 3d reconstruction in function space.
\newblock In {\em Proceedings of the IEEE Conference on Computer Vision and
  Pattern Recognition (CVPR)}, 2019.

\bibitem{nerf}
Ben Mildenhall, Pratul~P Srinivasan, Matthew Tancik, Jonathan~T Barron, Ravi
  Ramamoorthi, and Ren Ng.
\newblock Nerf: Representing scenes as neural radiance fields for view
  synthesis.
\newblock In {\em European conference on computer vision}, pages 405--421.
  Springer, 2020.

\bibitem{mirza2014conditional}
Mehdi Mirza and Simon Osindero.
\newblock Conditional generative adversarial nets.
\newblock {\em arXiv preprint arXiv:1411.1784}, 2014.

\bibitem{miyato2018spectral}
Takeru Miyato, Toshiki Kataoka, Masanori Koyama, and Yuichi Yoshida.
\newblock Spectral normalization for generative adversarial networks.
\newblock In {\em International Conference on Learning Representations (ICLR)},
  2018.

\bibitem{mustikovela2020self}
Siva~Karthik Mustikovela, Varun Jampani, Shalini~De Mello, Sifei Liu, Umar
  Iqbal, Carsten Rother, and Jan Kautz.
\newblock Self-supervised viewpoint learning from image collections.
\newblock In {\em Proceedings of the IEEE Conference on Computer Vision and
  Pattern Recognition (CVPR)}, 2020.

\bibitem{naseer2017deep}
Tayyab Naseer and Wolfram Burgard.
\newblock Deep regression for monocular camera-based 6-dof global localization
  in outdoor environments.
\newblock In {\em International Conference on Intelligent Robots and Systems
  (IROS)}, 2017.

\bibitem{nguyen2018rendernet}
Thu Nguyen-Phuoc, Chuan Li, Stephen Balaban, and Yong-Liang Yang.
\newblock Rendernet: A deep convolutional network for differentiable rendering
  from 3d shapes.
\newblock In {\em Neural Information Processing Systems (NeurIPS)}, 2018.

\bibitem{nguyen2019hologan}
Thu Nguyen-Phuoc, Chuan Li, Lucas Theis, Christian Richardt, and Yong-Liang
  Yang.
\newblock Hologan: Unsupervised learning of 3d representations from natural
  images.
\newblock In {\em The IEEE International Conference on Computer Vision (ICCV)},
  2019.

\bibitem{niemeyer2020giraffe}
Michael Niemeyer and Andreas Geiger.
\newblock Giraffe: Representing scenes as compositional generative neural
  feature fields.
\newblock In {\em Proceedings of the IEEE Conference on Computer Vision and
  Pattern Recognition (CVPR)}, 2021.

\bibitem{park2019deepsdf}
Jeong~Joon Park, Peter Florence, Julian Straub, Richard Newcombe, and Steven
  Lovegrove.
\newblock Deepsdf: Learning continuous signed distance functions for shape
  representation.
\newblock In {\em Proceedings of the IEEE Conference on Computer Vision and
  Pattern Recognition (CVPR)}, 2019.

\bibitem{peng2020convolutional}
Songyou Peng, Michael Niemeyer, Lars Mescheder, Marc Pollefeys, and Andreas
  Geiger.
\newblock Convolutional occupancy networks.
\newblock In {\em Proceedings of the European Conference on Computer Vision
  (ECCV)}, 2020.

\bibitem{philbin2007object}
James Philbin, Ondrej Chum, Michael Isard, Josef Sivic, and Andrew Zisserman.
\newblock Object retrieval with large vocabularies and fast spatial matching.
\newblock In {\em Proceedings of the IEEE Conference on Computer Vision and
  Pattern Recognition (CVPR)}, 2007.

\bibitem{radwan2018vlocnet++}
Noha Radwan, Abhinav Valada, and Wolfram Burgard.
\newblock Vlocnet++: Deep multitask learning for semantic visual localization
  and odometry.
\newblock {\em IEEE Robotics and Automation Letters}, 2018.

\bibitem{saito2019pifu}
Shunsuke Saito, Zeng Huang, Ryota Natsume, Shigeo Morishima, Angjoo Kanazawa,
  and Hao Li.
\newblock Pifu: Pixel-aligned implicit function for high-resolution clothed
  human digitization.
\newblock In {\em The IEEE International Conference on Computer Vision (ICCV)},
  2019.

\bibitem{schonberger2016structure}
Johannes~L Schonberger and Jan-Michael Frahm.
\newblock Structure-from-motion revisited.
\newblock In {\em Proceedings of the IEEE Conference on Computer Vision and
  Pattern Recognition (CVPR)}, 2016.

\bibitem{schwarz2020graf}
Katja Schwarz, Yiyi Liao, Michael Niemeyer, and Andreas Geiger.
\newblock Graf: Generative radiance fields for 3d-aware image synthesis.
\newblock In {\em Neural Information Processing Systems (NeurIPS)}, 2020.

\bibitem{shaham2019singan}
Tamar~Rott Shaham, Tali Dekel, and Tomer Michaeli.
\newblock Singan: Learning a generative model from a single natural image.
\newblock In {\em The IEEE International Conference on Computer Vision (ICCV)},
  2019.

\bibitem{sivic2003video}
Josef Sivic and Andrew Zisserman.
\newblock Video google: A text retrieval approach to object matching in videos.
\newblock In {\em The IEEE International Conference on Computer Vision (ICCV)},
  2003.

\bibitem{sun2021neural}
Guoxing Sun, Xin Chen, Yizhang Chen, Anqi Pang, Pei Lin, Yuheng Jiang, Lan Xu,
  Jingya Wang, and Jingyi Yu.
\newblock Neural free-viewpoint performance rendering under complexhuman-object
  interactions.
\newblock {\em arXiv preprint arXiv:2108.00362}, 2021.

\bibitem{tatarchenko2017octree}
Maxim Tatarchenko, Alexey Dosovitskiy, and Thomas Brox.
\newblock Octree generating networks: Efficient convolutional architectures for
  high-resolution 3d outputs.
\newblock In {\em The IEEE International Conference on Computer Vision (ICCV)},
  2017.

\bibitem{ulyanov2016instance}
Dmitry Ulyanov, Andrea Vedaldi, and Victor Lempitsky.
\newblock Instance normalization: The missing ingredient for fast stylization.
\newblock {\em arXiv preprint arXiv:1607.08022}, 2016.

\bibitem{valada2018deep}
Abhinav Valada, Noha Radwan, and Wolfram Burgard.
\newblock Deep auxiliary learning for visual localization and odometry.
\newblock In {\em IEEE International Conference on Robotics and Automation
  (ICRA)}, 2018.

\bibitem{walch2017image}
Florian Walch, Caner Hazirbas, Laura Leal-Taixe, Torsten Sattler, Sebastian
  Hilsenbeck, and Daniel Cremers.
\newblock Image-based localization using lstms for structured feature
  correlation.
\newblock In {\em The IEEE International Conference on Computer Vision (ICCV)},
  2017.

\bibitem{wang2021ibrnet}
Qianqian Wang, Zhicheng Wang, Kyle Genova, Pratul Srinivasan, Howard Zhou,
  Jonathan~T Barron, Ricardo Martin-Brualla, Noah Snavely, and Thomas
  Funkhouser.
\newblock Ibrnet: Learning multi-view image-based rendering.
\newblock In {\em Proceedings of the IEEE Conference on Computer Vision and
  Pattern Recognition (CVPR)}, 2021.

\bibitem{wang2004image}
Zhou Wang, Alan~C Bovik, Hamid~R Sheikh, and Eero~P Simoncelli.
\newblock Image quality assessment: from error visibility to structural
  similarity.
\newblock {\em IEEE TIP}, 2004.

\bibitem{wang2021nerf}
Zirui Wang, Shangzhe Wu, Weidi Xie, Min Chen, and Victor~Adrian Prisacariu.
\newblock Nerf $--$: Neural radiance fields without known camera parameters.
\newblock {\em arXiv preprint arXiv:2102.07064}, 2021.

\bibitem{wu2013towards}
Changchang Wu.
\newblock Towards linear-time incremental structure from motion.
\newblock In {\em 2013 International Conference on 3D Vision-3DV 2013}, 2013.

\bibitem{wu2017delving}
Jian Wu, Liwei Ma, and Xiaolin Hu.
\newblock Delving deeper into convolutional neural networks for camera
  relocalization.
\newblock In {\em IEEE International Conference on Robotics and Automation
  (ICRA)}, 2017.

\bibitem{wu20153d}
Zhirong Wu, Shuran Song, Aditya Khosla, Fisher Yu, Linguang Zhang, Xiaoou Tang,
  and Jianxiong Xiao.
\newblock 3d shapenets: A deep representation for volumetric shapes.
\newblock In {\em Proceedings of the IEEE Conference on Computer Vision and
  Pattern Recognition (CVPR)}, 2015.

\bibitem{yariv2020multiview}
Lior Yariv, Yoni Kasten, Dror Moran, Meirav Galun, Matan Atzmon, Basri Ronen,
  and Yaron Lipman.
\newblock Multiview neural surface reconstruction by disentangling geometry and
  appearance.
\newblock In {\em Neural Information Processing Systems (NeurIPS)}, 2020.

\bibitem{yen2020inerf}
Lin Yen-Chen, Pete Florence, Jonathan~T Barron, Alberto Rodriguez, Phillip
  Isola, and Tsung-Yi Lin.
\newblock inerf: Inverting neural radiance fields for pose estimation.
\newblock In {\em International Conference on Intelligent Robots and Systems
  (IROS)}, 2021.

\bibitem{zhang2020nerf++}
Kai Zhang, Gernot Riegler, Noah Snavely, and Vladlen Koltun.
\newblock Nerf++: Analyzing and improving neural radiance fields.
\newblock {\em arXiv preprint arXiv:2010.07492}, 2020.

\bibitem{zhang2018unreasonable}
Richard Zhang, Phillip Isola, Alexei~A Efros, Eli Shechtman, and Oliver Wang.
\newblock The unreasonable effectiveness of deep features as a perceptual
  metric.
\newblock In {\em Proceedings of the IEEE Conference on Computer Vision and
  Pattern Recognition (CVPR)}, 2018.

\bibitem{zhang2018tutorial}
Zichao Zhang and Davide Scaramuzza.
\newblock A tutorial on quantitative trajectory evaluation for visual
  (-inertial) odometry.
\newblock In {\em International Conference on Intelligent Robots and Systems
  (IROS)}, 2018.

\bibitem{zhou2019continuity}
Yi Zhou, Connelly Barnes, Jingwan Lu, Jimei Yang, and Hao Li.
\newblock On the continuity of rotation representations in neural networks.
\newblock In {\em Proceedings of the IEEE Conference on Computer Vision and
  Pattern Recognition (CVPR)}, 2019.

\end{thebibliography}
}

\clearpage
\appendix
\section{Pose Distribution Analysis}
We analyze mismatch of pose distribution on the Synthetic-NeRF dataset in Tab.~\ref{pose_dis_tab}. We change the range of pose sampling space and report the novel view synthesis quality on the chair scene in the Synthetic-NeRF dataset. The pose sampling space is represented by camera position and camera rotation. The camera position is determined by three parameters: radius, elevation, and azimuth. The camera rotation is calculated from the camera position, camera lookat points, and camera up vector. As is shown in the table, slightly changing the radius or lookat points does not depreciate the performance greatly. For the third pose distribution of changing the elevation range from $[0, 90]$ to $[-90, 90]$, we note that this setting will generate views that are not seen by any training image which will disturb the inversion network. It demonstrates that our method relies on a reasonable camera sampling space but not necessarily an accurate one. 

\section{Additional Results} 
\myparagraph{Image Size.} We provide the test on how the input image size influences the COLMAP on the Synthetic-NeRF dataset. The input images are 108 images with the size of $400\times400$ or $800\times800$, we report the registered images concerning image size in Tab.~\ref{image_size}. As illustrated, the COLMAP is sensitive to the image size. It works satisfactorily with the original image size $800\times800$ but struggles to register much more images on the downsampled images. For this reason, we provide only results of COLMAP-based NeRF with registered poses from images with the size of $800\times800$.

\section{Applications}

\begin{figure}[t]
	\begin{center}
		\includegraphics[width=1.0\linewidth]{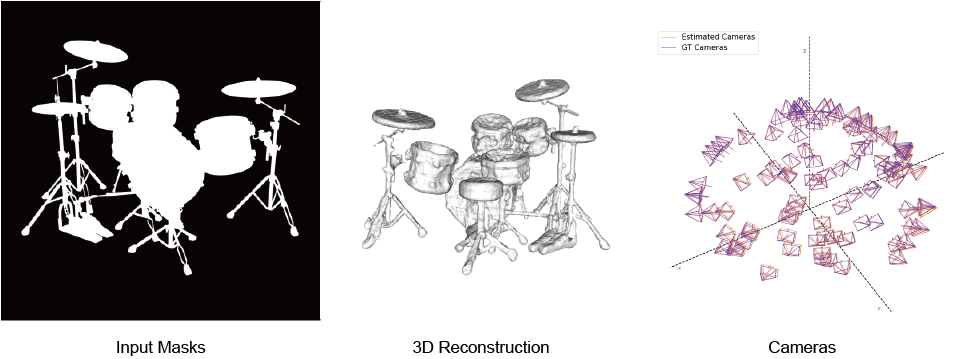}
	\end{center}
	\caption{\textbf{3D reconstruction and camera pose estimation from a collection of masks without pose information.} }
	\label{fig:application_1}
\end{figure}

\myparagraph{3D Reconstruction from Unposed Masks} In Fig.~\ref{fig:application_1}, we learn 3D representation and camera poses from a collection of unposed masks by optimizing the radiance fields and camera poses simultaneously. Specifically, we treat the mask as a 1-channel image and render it with volume rendering as RGB images. With the trained NeRF model, we then extract the 3D representation using the marching cubes algorithm~\cite{lorensen1987marching} following the original NeRF script~\footnote{https://github.com/kwea123/nerf\_pl}. This case further demonstrates that our architecture can estimate camera poses from high-level features of an image without reliance on keypoints or textures, which is fundamentally different from conventional pose estimation methods. This ability of our method can be applied to other applications, such as the task of reconstructing transparent objects whose visual appearance is too complex for image-based reconstruction. Since it is much easier to obtain the masks of their shapes either by semantic segmentation tools or other sensors.

\begin{figure}[t]
	\begin{center}
		\includegraphics[width=1.0\linewidth]{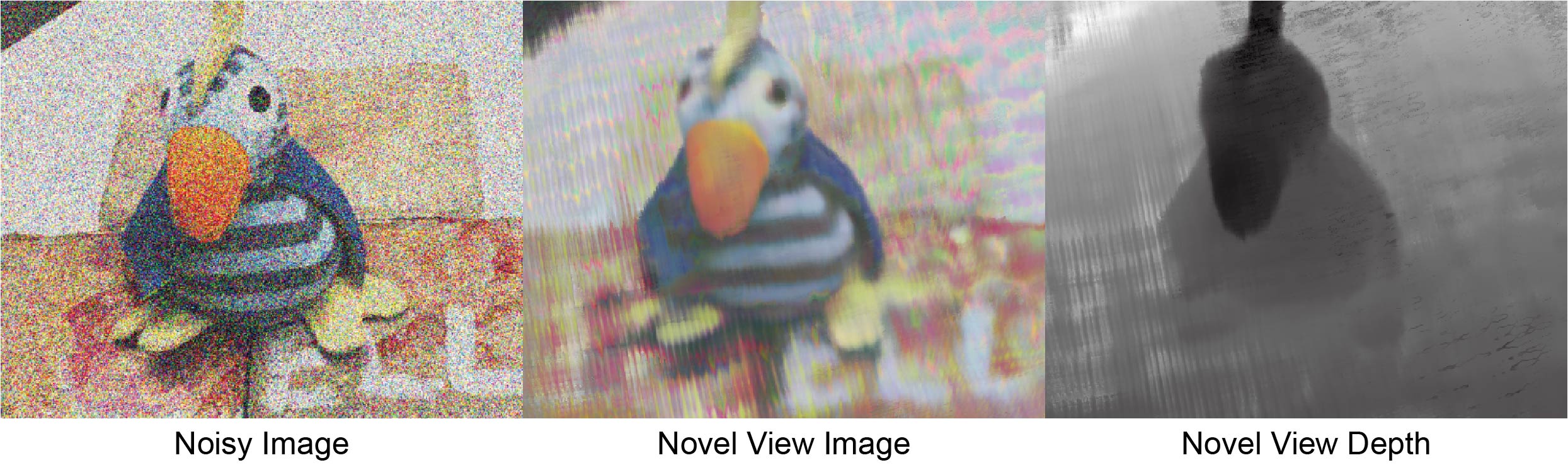}
	\end{center}
	\caption{\textbf{Image Noise Analysis.} Despite adding intense noise on training images, our method is able to learn accurate radiance fields and camera poses of the noisy images while COLMAP-based NeRF methods completely fail.}
	\label{fig:image_noise_vis}
\end{figure}

\myparagraph{Image Noise Analysis} In Fig~\ref{fig:image_noise_vis}, we test our method on images with intense noise. The COLMAP-based NeRF methods completely fail to estimate the camera poses of images with Gaussian noise $\mathcal{N}(0, 0.5^2)$, leading to failure of learning the radiance fields. In contrast, our method is not sensitive to noise and still able to render novel view and depth map with less noise. We demonstrate the accuracy of pose estimation qualitatively by a rendered image (the middle image) with the estimated pose of the left noisy image. 

\begin{table}[t]
	\centering
	\begin{tabular}{*{5}{c}}
		\toprule
		\multicolumn{1}{c}{Radius} & \multicolumn{1}{c}{Ele(deg)} & \multicolumn{1}{c}{Azi(deg)} & 
		\multicolumn{1}{c}{Lookat} & \multicolumn{1}{c}{PSNR} \\
		
		\midrule
		$4$ & $\left[0, 90\right]$ & $\left[0, 360\right]$ & (0, 0, 0) & 31.30 \\
		$\left[3, 5\right]$ & $\left[0, 90\right]$ & $\left[0, 360\right]$ & (0, 0, 0) & 29.68 \\
		$4$ & $\left[-90, 90\right]$ & $\left[0, 360\right]$ & (0, 0, 0) & 23.08 \\
		$4$ & $\left[0, 90\right]$ & $\left[0, 360\right]$ & $\mathcal{N}(0, 0.01^2)$ & 30.48 \\
		\bottomrule
	\end{tabular}
	\caption{\textbf{Pose Distribution Analysis.} We change the camera sampling space by singly change Radius, Elevation(Ele), Azimuth(Azi), and lookat point, and report the novel view synthesis quality on the chair scene in Synthetic-NeRF dataset.}
	\label{pose_dis_tab}
\end{table}

\begin{table}[t]
	\centering
	\begin{tabular}{*{7}{c}}
		\toprule
		\multicolumn{1}{c}{Size} &
		\multicolumn{1}{c}{Chair} & \multicolumn{1}{c}{Drums} & \multicolumn{1}{c}{Hotdog} & 
		\multicolumn{1}{c}{Lego} &
		\multicolumn{1}{c}{Mic} &
		\multicolumn{1}{c}{Ship}
		\\
		
		\midrule 
		400 & 108 & 87 & 98 & 108  & 19 & 82 \\
		800 & 108 & 100 & 104 & 108 & 84 & 108 \\
		\bottomrule
	\end{tabular}
	\caption{\textbf{Image numbers Analysis.} We reduce the training image numbers and compare with the COLMAP-based NeRF. }
	\label{image_size}
\end{table}

\end{document}